# SISTEM PENDUKUNG KEPUTUSAN KELAYAKAN TKI KE LUAR NEGERI MENGGUNAKAN FMADM

**Ardina Ariani[1)], Leon Andretti Abdillah[2*)] Firamon Syakti[3)]**

[1,2)] Program Studi Sistem Informasi, Fakultas Ilmu Komputer, Universitas Bina Darma
[3)] Program Studi Komputerisasi Akuntansi, Fakultas Ilmu Komputer, Universitas Bina Darma
Jl. Ahmad Yani No.12, Palembang, 30264
Telp : (0711) 515679, Fax : (0711) 515581
E-mail : leon.abdillah@yahoo.com[2*)]

*Abstract*

*BP3TKI Palembang is the government agencies that coordinate, execute and selection of prospective migrants registration and placement. To simplify the existing procedures and improve decision-making is necessary to build a decision support system (DSS) to determine eligibility for employment abroad by applying Fuzzy Multiple Attribute Decision Making (FMADM), using the linear sequential systems development methods. The system is built using Microsoft Visual Basic. Net 2010 and SQL Server 2008 database. The design of the system using use case diagrams and class diagrams to identify the needs of users and systems as well as systems implementation guidelines. Decision Support System which is capable of ranking the dihasialkan to prospective migrants, making it easier for parties to take keputusna BP3TKI the workers who will be flown out of the country.*

*Abstrak*

*BP3TKI Palembang adalah instansi pemerintah yang mengoordinasikan, melaksanakan pendaftaran dan seleksi calon TKI serta lokasi penempatannya. Untuk memudahkan prosedur yang ada dan meningkatkan pengambilan keputusan, perlu dibangun sebuah sistem pendukung keputusan (SPK). SPK ini bertujuan untuk menentukan kelayakan tenaga kerja ke luar negeri dengan menerapkan Fuzzy Multiple Attribute Decision Making (FMADM) yang menggunakan metode pengembangan sistem linier sequential. Sistem ini dibangun dengan menggunakan Microsoft Visual Basic. Net 2010 dan database SQL Server 2008. Perancangan sistem menggunakan use case diagram dan class diagram untuk mengidentifikasi kebutuhan pengguna dan sistem serta pedoman implementasi sistem. Sistem Penunjang Keputusan yang dihasilkan mampu melakukan perangkingan terhadap calon TKI, sehingga memudahkan pihak BP3TKI dalam mengambil keputusan mengenai TKI yang akan diberangkatkan ke luar negeri.*

**Kata kunci:** *sistem pendukung keputusan, kelayakan, TKI, FMADM*

## 1. PENDAHULUAN

Indonesia merupakan Negara yang padat penduduk, namun belum memiliki lapangan pekerjaan yang cukup untuk peningkatan taraf hidup penduduk. Hal ini mendorong banyak penduduk yang menjadi tenaga kerja Indonesia ke luar negeri. Namun tenaga kerja Indonesia yang layak untuk dipekerjakan di luar negeri harus memiliki kriteria khusus, yaitu: usia, pendidikan, keterampilan, pengetahuan, dan pengalaman kerja. Kriteria-kriteria tersebut menjadi acuan dalam proses penyeleksian kelayakan tenaga kerja Indonesia ke luar negeri.

Dalam Undang-Undang Republik Indonesia No.39 Tahun 2004 tentang Penempatan dan Perlindungan Tenaga Kerja Indonesia di Luar Negeri, TKI adalah setiap warga negara Indonesia yang memenuhi syarat untuk bekerja di luar negeri dalam hubungan kerja untuk jangka waktu tertentu dengan menerima upah. Undang-undang ini juga menerangkan bahwa penentuan kelayakan tenaga kerja atau seleksi tenaga kerja adalah proses pencarian karyawan untuk menyeleksi calon tenga kerja yang dianggap memenuhi kriteria yang sesuai dengan karakter pekerjaan yang dilamar. Kelayakan tenaga kerja Indonesia ke luar negeri ditentukan berdasarkan kriteria: usia, keterampilan, pengetahuan, pengalaman kerja, dan pendidikan.

Balai Pelayanan Penempatan dan Perlindungan Tenaga Kerja Indonesia yang disebut BP3TKI adalah Unit Pelaksana Teknis BNP2TKI yang bertugas mengkoordinasikan, melaksanakan pendaftaran dan seleksi calon TKI serta penempatannya. Selama ini dalam pengambilan keputusan penentuan kelayakan TKI yang akan diberangkatkan ke luar negeri pada tahap seleksi calon TKI di BP3TKI dilakukan dengan penyeleksian kelengkapan syarat dokumen yang harus dipenuhi oleh para calon TKI. Prosedur yang berjalan sampai pada saat ini, secara umum telah dapat menentukan kelayakan calon TKI diberangkatkan bekerja ke luar negeri. Namun prosedur tersebut perlu diadakan peningkatan dalam segi kualitas pengambilan keputusan dengan didukung oleh Sistem Pendukung Keputusan dalam pengambilan keputusan penentuan kelayakan tenaga kerja ke luar negeri.

Mengingat pentingnya sutau sistem untuk membantu pengambilan keputusan akan kelayakan TKI ke luar negeri, maka penulis tertarik untuk melakukan penelitian pengembangan Sistem Pendukung Keputusan dalam menentukan kelayakan tenaga kerja ke luar negeri dengan menggunakan *Fuzzy Multiple Attribute Decision Making* (FMADM).

Beberapa metode yang sering digunakan dalam pemodelan Sistem Pendukung Keputusan (SPK) antara lain: AHP (*Analytical Hierarkhi Process*), TOPSIS (*Technique for Order Preference by Similarity to Ideal Solution*), dan FMADM (*Fuzzy Multiple Attribute Decision Making*). Penulis memilih metode FMADM untuk digunakan dalam penelitian ini dimana langkah penyeleksian alternatifnya lebih pendek namun akan tetap menghasilkan keputusan optimal dalam menentukan alternatif terbaik dari berbagai alternatif berdasarkan kriteria tertentu.

Pada penelitian sebelumnya metode pengambilan keputusan telah banyak digunakan untuk berbagai penelitian. Gerdon (2011) melakukan penelitian untuk menentukan penerima beasiswa berdasarkan lima kriteria, yaitu: nilai IPK, penghasilan orang tua, semester, jumlah tanggungan orang tua, dan usia. Gerdon menggunakan metode SAW (*Simple Additive Weighting*) digunakan untuk melakukan perhitungan dalam metode FMADM. Penelitian dilakukan dengan menentukan nilai bobot untuk setiap kriteria, kemudian dilakukan proses perangkingan sehingga mendapatkan alternatif optimal yaitu mahasiswa terbaik yang akan dipertimbangkan oleh pengambil keputusan untuk memperoleh beasiswa. Selanjutnya, Wardhani (2012) dalam penelitiannya "Seleksi *Supplier* Bahan Baku dengan Metode TOPSIS FMADM" melakukan pemilihan *supplier* dengan mengukur kinerja *supplier* pada objek penelitian *supplier* bahan baku seperti semen, pasir, besi, batu pondasi dan kayu.

## 2. METODOLOGI

Sistem Pendukung Keputusan yang dibangun menggunakan pendekatan *linear sequential model* (Pressman, 2011), sedangkan teknik analisis keputusan akan digunakan metode FMADM.

### 2.1 Analisis Sistem Berjalan

Sampai pada saat ini dalam pengambilan keputusan kelayakan TKI yang akan diberangkatkan ke luar negeri pada tahap seleksi calon TKI di BP3TKI dilakukan dengan penyeleksian kelengkapan calon TKI. Prosedur yang berjalan sampai pada saat ini, secara umum telah dapat menentukan kelayakan calon TKI yang akan diberangkatkan bekerja ke luar negeri. Namun prosedur tersebut perlu diadakan peningkatan dalam segi kualitas pengambilan keputusan dengan didukung oleh Sistem Pendukung Keputusan dalam pengambilan keputusan kelayakan tenaga kerja ke luar negeri dengan menerapkan logika FMADM di dalam proses penyeleksiannya.

### 2.2 Analisis Kebutuhan Input dan Output

Kebutuhan *input* pada pembangunan sistem ini adalah semua objek yang dibutuhkan oleh sistem yang dibangun yaitu sistem pendukung keputusan kelayakan tenaga kerja ke luar negeri dalam menghasilkan informasi (*output*) untuk membantu pengambilan keputusan untuk menentukan kelayakan tenaga kerja ke luar negeri, objek tersebut yaitu: data CTKI (calon tenaga kerja Indonesia), data PPTKIS (Pelaksana Penempatan Tenaga Kerja Indonesia Swasta), data negara penempatan dan data kriteria penyeleksian.

Kebutuhan *output* atau keluaran dari sistem yang dibangun adalah semua keluaran yang berupa informasi yang dihasilkan dari sistem pendukung keputusan kelayakan tenaga kerja ke luar negeri. Informasi atau keluaran dari sistem yang dibangun adalah informasi mengenai alternatif terpilih dari sejumlah alternatif dari hasil penyeleksian dengan memberikan urutan perangkingan dari tertinggi hingga terendah.

### 2.3 Analisis Kriteria Penyeleksian

Dalam penelitian ini, proses penyeleksian calon TKI menggunakan metode *Fuzzy Multi-Attribute Decision Making* membutuhkan beberapa kriteria, terdapat 4 kriteria yang digunakan yaitu: C1 = Usia, C2 = Pendidikan, C3 = Psikotes, dan C4 = Pengalaman kerja. Kriteria-kriteria ini dipilih berdasarkan kriteria yang memang telah digunakan oleh Badan Nasional Penempatan dan Perlindungan Tenaga

Kerja Nasional (BP3TKI) dalam penyeleksian calon TKI. Kriteria-kriteria tersebut selanjutnya akan dijadikan *input* dalam langkah-langkah proses penyeleksian dengan metode FMADM.

**2.4 Analisis Logika Proses**

Adapun tahapan proses logika dengan metode ini adalah sebagai berikut:

a)  Menentukan kriteria dan bobot keputusan. Bobot keputusan menunjukkan kepentingan relatif dari setiap kriteria, W = (w1, w2,…, wn). Pada FMADM akan dicari bobot kepentingan dari setiap kriteria. Bobot untuk masing-masing kriteria tersebut yaitu (W) = Cukup penting, Penting, Sangat penting, Penting.
b)  Mengubah bobot kriteria dalam bentuk bilangan *fuzzy* Dari bobot kriteria tersebut diubah dalam bentuk bilangan *fuzzy*. Bilangan *fuzzy* dari bobot kriteria tersebut yaitu tidak penting (TP), cukup penting (CP), penting (P), dan sangat penting (SP). Bilangan *fuzzy* tersebut dikonversi ke bilangan *crisp* tergambar pada gambar 1.
c)  Setiap kriteria/atribut dikonversikan ke bilangan *crisp*.
d)  Memberikan nilai alternatif pada setiap kriteria. Setiap alternatif yang menjadi peserta seleksi diberikan nilai berdasarkan empat kriteria.
e)  Melakukan perhitungan normalisasi matriks (R). Setiap nilai yang dimiliki alternatif pada setiap kriteria dibagi dengan nilai maksimum per kriteria.

**2.5 Use Case Diagram**

Diagram *use case* telah diusulkan dalam UML sebagai notasi untuk menggambarkan persyaratan perangkat lunak sistem dan perilaku (Shen, 2003). Pada penelitian ini, *Use Case Diagram* menunjukkan terdapat 2 aktor yang saling berhubungan, yaitu calon TKI dan staf bagian penempatan.

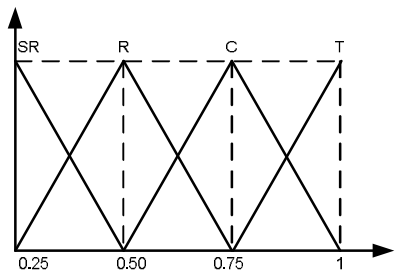

*Gambar 1. Bilangan fuzzy untuk bobot kriteria*

Tabel 1. Kriteria 1 (C1) Usia

| Usia (X) | Bilangan *Crisp* |
|---|---|
| $X \geq 18 \leq 20$ | 1 |
| $X \geq 21 \leq 23$ | 0.75 |
| $X \geq 24 \leq 26$ | 0.50 |
| $X \geq 27 \leq 30$ | 0.25 |
| $X \geq 31 \leq 35$ | 0 |

Tabel 2. Kriteria 2 (C2) Pendidikan

| Pendidikan | Bilangan *Crisp* |
|---|---|
| SMP | 0 |
| SMA | 0.25 |
| DI-DIII | 0.50 |
| DIV | 0.75 |
| S1 | 1 |

Tabel 3. Kriteria 3 (C3) Psikotes

| Psikotes | Bilangan *Crisp* |
|---|---|
| Disarankan | 1 |
| BelumDisarankan | 0 |

Tabel 4. Kriteria 4 (C4) Pengalaman Kerja

| Pengalaman Kerja (Y) | Bilangan *Crisp* |
|---|---|
| Y = 0 Tahun | 0 |
| $Y \geq 1 \leq 3$ Tahun | 0.25 |
| $Y \geq 4 \leq 6$ Tahun | 0.50 |
| $Y \geq 7 \leq 9$ Tahun | 0.75 |
| $Y \geq 10$ Tahun | 1 |

Aktor calon TKI menyerahkan berkas yang dibutuhkan dalam penyeleksian lalu aktor staf bagian penempatan login agar dapat berinteraksi dengan sistem untuk memasukkan data calon TKI, nilai kriteria, melakukan penyeleksian dengan *fuzzy*, dan membuat laporan hasil penyeleksian.

**2.6 Class Diagram**

Sebuah *class diagram* adalah yang paling mendasar dan banyak digunakan diagram UML. UML ini menunjukkan pandangan statis dari sebuah sistem, yang terdiri dari kelas, antar hubungan mereka (termasuk generalisasi, spesialisasi, asosiasi, agregasi dan komposisi), operasi dan atribut dari kelas (Szlenk, 2006).

Diagram ini juga memberikan alur hubungan antar kelas yang satu dengan kelas lainnya. Terdapat lima kelas yaitu ctki, *input* nilai awal, bilangan crisp, normalisasi matriks, perangkingan dan hasil seleksi.

Ariani, dkk., Sistem Pendukung Keputusan Kelayanan TKI ke Luar Negeri Menggunakan FMADM

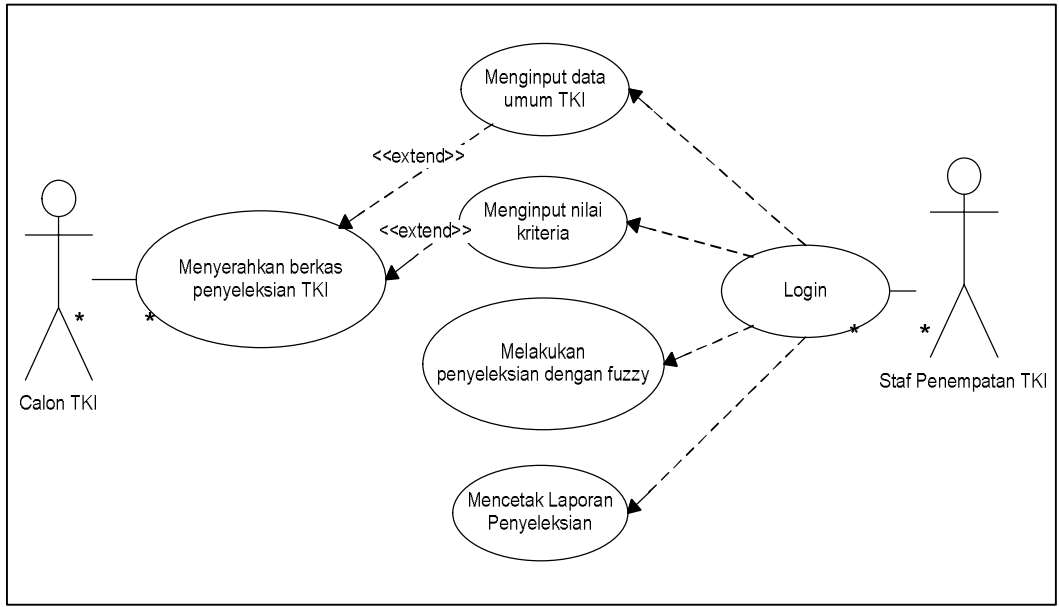

*Gambar 2. Use Case Diagram*

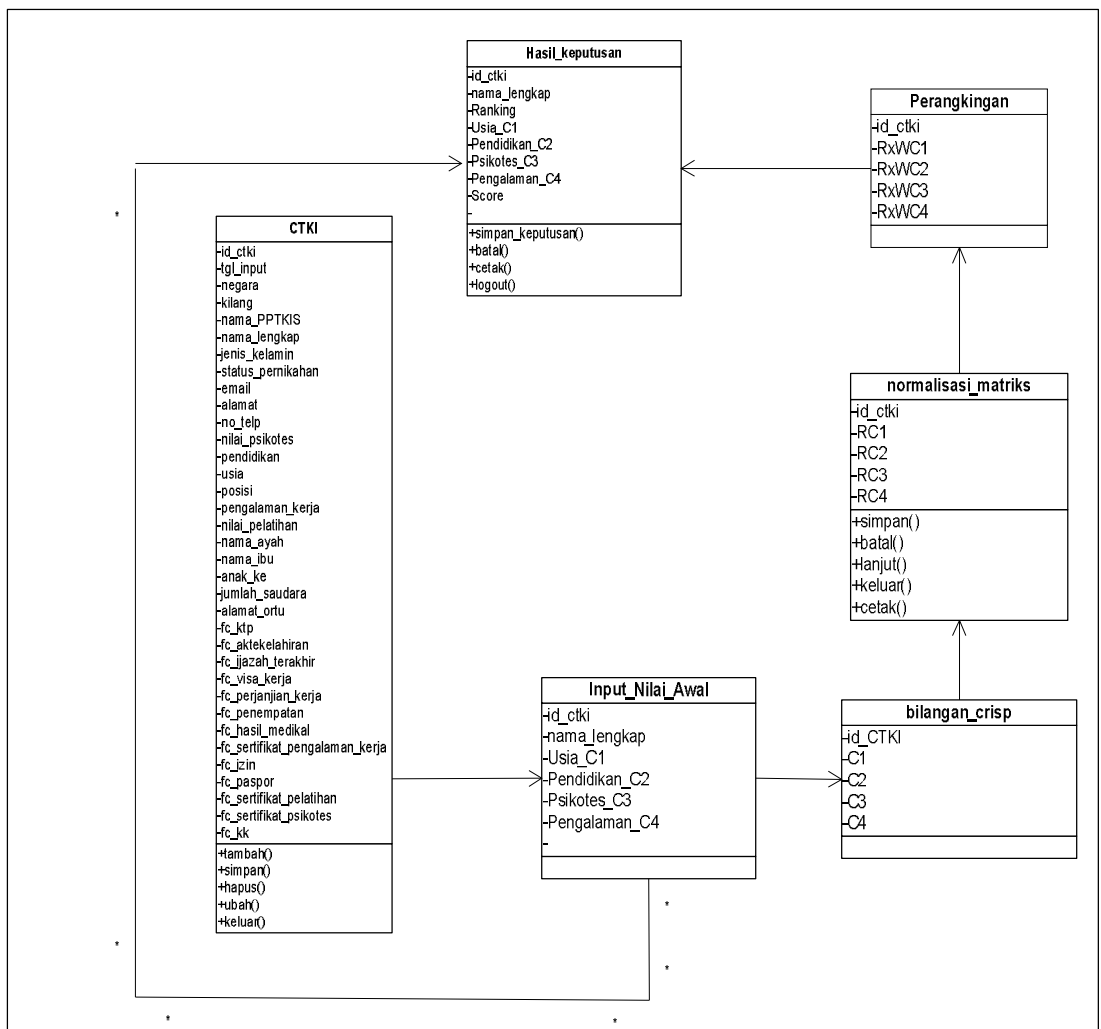

*Gambar 4. Class Diagram*

**2.7** *Fuzzy Multi-Attribute Decision Making* **(FMADM)**

FMADM adalah suatu metode yang digunakan untuk mencari alternatif optimal dari sejumlah alternatif dengan kriteria tertentu. Inti dari FMADM adalah menentukan nilai bobot untuk setiap atribut, kemudian dilanjutkan dengan proses perangkingan yang akan menyeleksi alternatif yang sudah diberikan. (Kusumadewi, dkk, 2006).

Algoritma FMADM adalah: 1) Memberikan nilai setiap alternatif (Ai) pada setiap kriteria (Cj) yang sudah ditentukan, dimana nilai tersebut diperoleh berdasarkan nilai crisp; i=1,2,...m dan j=1,2,...n., 2) Memberikan nilai bobot (W) yang juga didapatkan berdasarkan nilai crisp, 3) Melakukan normalisasi matriks dengan cara menghitung nilai rating kinerja ternormalisasi (rij) dari alternatif Ai pada atribut Cj berdasarkan persamaan yang disesuaikan dengan jenis atribut (atribut keuntungan = MAKSIMUM atau atribut biaya = MINIMUM). Apabila berupa atribut keuntungan maka nilai crisp (Xij) dari setiap kolom atribut dibagi dengan nilai crisp MAX (MAX Xij) dari tiap kolom, sedangkan untuk atribut biaya, nilai crisp MIN (MIN Xij) dari tiap kolom atribut dibagi dengan nilai crisp (Xij) setiap kolom, dan 4) Melakukan proses perangkingan dengan cara mengalikan matriks ternormalisasi (R) dengan nilai bobot (W).

Menentukan nilai preferensi untuk setiap alternatif (Vi) dengan cara menjumlahkan hasil kali antara matriks ternormalisasi (R) dengan nilai bobot (W). Nilai Vi yang lebih besar mengindikasikan bahwa alternatif Ai lebih terpilih.

**3. HASIL dan PEMBAHASAN**

Berdasarkan hasil penelitian yang telah dilakukan pada BNPPTKI maka didapatkan hasil akhir sebuah sistem yaitu Sistem Pendukung Keputusan Kelayakan Tenaga Kerja ke Luar Negeri dengan menggunakan FMADM.

**3.1 Data Calon TKI**

*Form input* data CTKI merupakan *form* yang digunakan untuk menginput data calon tenaga kerja Indonesia yang akan diseleksi. Pada *form* ini terdapat 5 tombol yang digunakan untuk menambah, menyimpan, mengubah, menghapus data, dan menutup *form*.

**3.2 Normalisasi Matriks**

*Form* normalisasi matriks adalah *form* yang digunakan untuk proses konversi data awal ke bilangan *crisp* dan proses normalisasi matriks. *Form* ini terdiri dari 3 *datagridview* yang berguna untuk menampilkan nilai awal yang diambil dari kriteria yang didapat dari data calon TKI, *datagridview* bilangan crisp digunakan untuk menampilkan nilai hasil konversi nilai awal ke bilangan *crisp* berdasarkan kriteria, dan *datagridview* konversi normalisasi matriks untuk menampilkan hasil perhitungan normalisasi matriks terhadap bilangan *crisp* untuk setiap kriteria di berbagai alternatif.

**3.3 Bobot Kriteria**

*Form* bobot kriteria merupakan *form* yang digunakan untuk menampilkan tabel bobot kriteria. *Form* ini digunakan untuk membantu *user* atau pengguna untuk melihat berapa nilai bobot untuk setiap kriteria dalam penyeleksian setiap alternatif.

*Gambar 13. Form Input Data CTKI*

*Gambar 14. Form Normalisasi Matriks*

*Gambar 15. Form Bobot Kriteria*

*Gambar 16. Form Perangkingan*

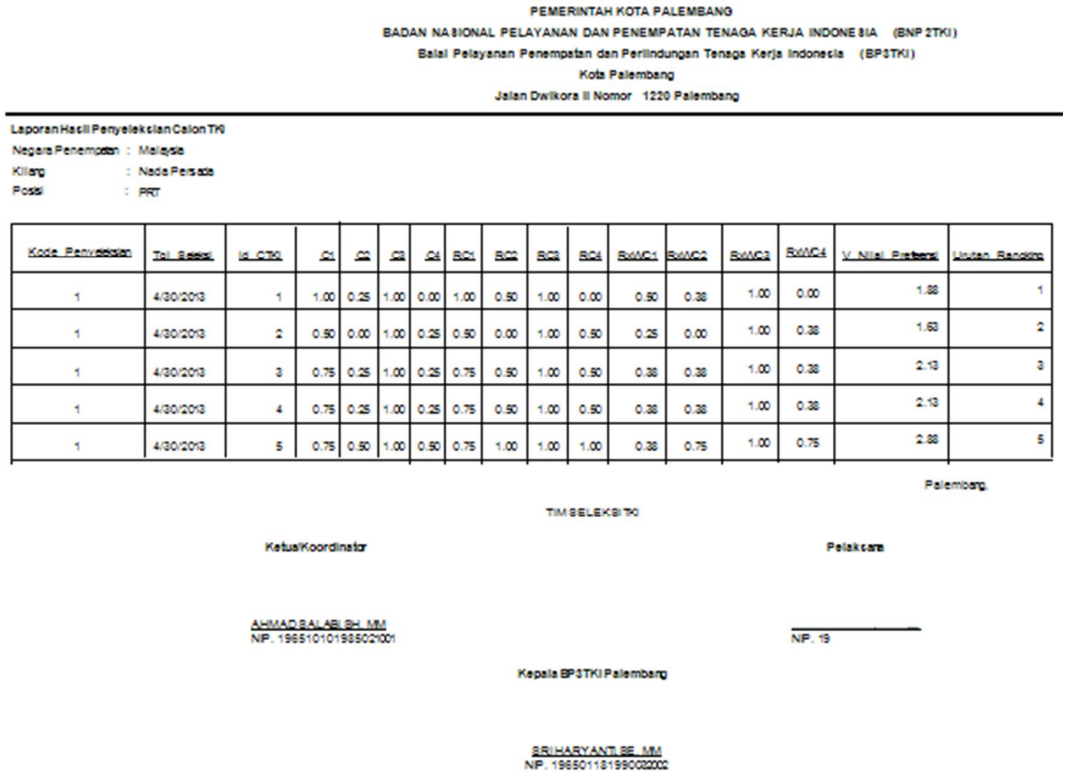

*Gambar 17. Laporan Hasil Penyeleksian*

### 3.4 Perangkingan

*Form* ini menampilkan hasil perangkingan dan mengurutkan perangkingan berdasarkan nilai preferensi setiap alternatif. Nilai pada *form* ini didapatkan dari *form* normalisasi pada *datagridview*, nilai normalisasi selanjutnya dikalikan dengan bobot kriteria yang menghasilkan nilai preferensi (V), dan mengurutkannya berdasarkan besarnya nilai preferensi sebagai rangking calon tenaga kerja Indonesia.

### 3.5 Laporan

Laporan ini berisi nilai yang dihasilkan pada proses penyeleksian, terdapat nilai untuk bilangan *crisp*, normalisasi matriks setiap kriteria pada setiap alternatif, dan besarnya nilai preferensi perangkingan untuk setiap alternatif (calon TKI).

### 4. SIMPULAN dan SARAN

Berdasarkan penelitian yang telah dilakukan dengan tema Sistem Pendukung Kelayakan Tenaga Kerja ke Luar Negeri dengan menggunakan metode *Fuzzy Multiple Attribute Decision Making* (FMADM) pada BP3TKI Kota Palembang, maka penulis mengambil beberapa kesimpulan yaitu:

1. Sistem Penunjang Keputusan yang dibangun mampu menghasilkan daftar rangking calon TKI, sehingga memudahkan pihak BP3TKI dalam mengambil keputusan TKI yang layak kirim kerja ke luar negeri.

2. Sistem ini dapat menerima *input* data calon TKI, melakukan konversi data calon TKI berdasarkan kriteria: usia, pendidikan terakhir, psikotes dan lama pengalaman kerja ke bilangan *crisp*, melakukan perhitungan normalisasi dan perangkingan, sehingga dapat menghasilkan informasi hasil penyeleksian kelayakan tenaga kerja ke luar negeri berdasarkan urutan rangking.

3. Sistem Pendukung Kelayakan Tenaga Kerja ke Luar Negeri dengan menggunakan metode *Fuzzy Multiple Attribute Decision Making* (FMADM) pada BP3TKI Kota Palembang dapat memberikan validasi data agar terhindar dari pengulangan penyimpanan data yang sama.

Dari kesimpulan yang telah dikemukakan, beberapa saran yang dapat diterapkan untuk penelitian berikutnya. Untuk pengembangan selanjutnya sistem dapat berupa website. Jika sistem ini dapat diterapkan di Badan Nasional Penempatan dan Perlidungan Tenaga Kerja Indonesia (BP3TKI) diharapkan agar dilakukan pelatihan bagi staf yang yang diberi kewenangan dalam menjalankan sistem ini. Evaluasi yang rutin pada BPTKI sendiri dapat membantu sistem berjalan dengan optimal.

## 5. DAFTAR RUJUKAN